\documentclass[sigconf,nonacm]{acmart}
\AtBeginDocument{%
  }

\usepackage{tabularx}  
\usepackage{booktabs}  
\usepackage{multirow}  
\usepackage{ragged2e}  
\usepackage{array}  
\usepackage{enumitem }
\usepackage{comment}

\begin{document}

\title{AgentReputation: A Decentralized Agentic AI Reputation Framework}

\author{Mohd Sameen Chishti}
\affiliation{%
  \institution{Department of Computer Science \\ NTNU}
  \city{Trondheim}
  \country{Norway}
}
\email{mohd.s.chishti@ntnu.no}
\orcid{0000-0003-3977-8488}

\author{Damilare Peter Oyinloye}
\affiliation{%
  \institution{Department of Computer Science \\ NTNU}
  \city{Trondheim}
  \country{Norway}
}
\email{peter.d.oyinloye@ntnu.no}
\orcid{0000-0002-4925-5042}

\author{Jingyue Li}
\affiliation{%
  \institution{Department of Computer Science \\ NTNU}
  \city{Trondheim}
  \country{Norway}
}
\email{jingyue.li@ntnu.no}
\orcid{0000-0000-0000-0000}




\renewcommand{\shortauthors}{Chishti et al.}

\begin{abstract}
Decentralized, agentic AI marketplaces are rapidly emerging to support software engineering tasks such as debugging, patch generation, and security auditing, often operating without centralized oversight. However, existing reputation mechanisms fail in this setting for three fundamental reasons: agents can strategically optimize against evaluation procedures; demonstrated competence does not reliably transfer across heterogeneous task contexts; and verification rigor varies widely, from lightweight automated checks to costly expert review. Current approaches to reputation drawing on federated learning, blockchain-based AI platforms, and large language model safety research are unable to address these challenges in combination. We therefore propose \textbf{AgentReputation}, a decentralized, three-layer reputation framework for agentic AI systems. The framework separates task execution, reputation services, and tamper-proof persistence to both leverage their respective strengths and enable independent evolution. The framework introduces explicit verification regimes linked to agent reputation metadata, as well as context-conditioned reputation cards that prevent reputation conflation across domains and task types. In addition, AgentReputation provides a decision-facing policy engine that supports resource allocation, access control, and adaptive verification escalation based on risk and uncertainty. Building on this framework, we outline several future research directions, including the development of verification ontologies, methods for quantifying verification strength, privacy-preserving evidence mechanisms, cold-start reputation bootstrapping, and defenses against adversarial manipulation. 
 
\end{abstract}



\maketitle

\section{Introduction}

Recent advancements in large language models (LLMs) are changing how software engineering (SE) processes shift from human-only workflows to human-agent pipelines \cite{10.1145/3597503.3649399,10.1145/3695988}. The increasing use of tool-using agents can now autonomously navigate repositories, localize faults, run tests, and produce patches \cite{10449667,10.1145/3650212.3680384,11121619}. At the same time, software ecosystems are increasingly open-ended, allowing agents to contribute artifacts that affect the software supply chain through patches, dependency choices, and security \cite{10.1145/3708531,10.1145/3650212.3680397}. The agents will contribute relevant changes to the development life cycle without any centralized oversight \cite{10962241, hu2025trustlessautonomyunderstandingmotivations}. This shift is expected to improve the efficiency of software engineering processes and support higher throughput as adoption grows \cite{10.1007/978-3-031-72781-8_2}, but introduces a trust trilemma. In an environment where agents are strategic, tasks are heterogeneous, and verification quality varies significantly, how can task owners reliably assess agent competence?

The traditional reputation system, which often aggregates historical ratings or credentials \cite{10.1145/3362168,10.1108/IJWIS-12-2023-0247, huang2025novelzerotrustidentityframework}, is ill-suited for SE workflows as the artifacts at stake directly impact the correctness, maintainability, and security of software. The AI agents have strategic capabilities that learn deceptive behaviors, allowing them to optimize for approval metrics rather than genuine competence \cite{curvo2025traitorsdeceptiontrustmultiagent}.  The agents operate across domains with varied requirements, like proficiency in debugging or security auditing.  A monolithic reputation score dangerously conflates such contexts. Also, the verification itself ranges from superficial automated checks to rigorous expert review, yet the current system treats all validation as equally informative \cite{Pysmennyi_Kyslyi_Kleshch_2025}.

Existing research on reputation addresses only fragments of the trilemma. Federated learning assumes collaborative honesty \cite{8994206,8733825,9488743}, a blockchain-based AI platform lacks contextual granularity \cite{7973733, mcconaghy2022ocean}, and LLM safety research aims to characterize threats without providing operational defenses \cite{park2024ai,hubinger2024sleeperagentstrainingdeceptive,xu2025hallucinationinevitableinnatelimitation}. The missing piece here is a unified vision for reputation as active infrastructure. An infrastructure that can ground trust in verifiable evidence, preserve contextual integrity, and actively govern operational and economic decisions.

This paper attempts to articulate the vision of reputation as active infrastructure. We argue that there is a need to move beyond scalar scores and must design a framework that is: \emph{Evidence-Based}, providing explicit verification regimes with quantified strength, \emph{Contextual},  preventing the cross-domain aggregation through structured and task-specific assessment, and \emph{Decision-facing}, which can dynamically govern allocation, access rights, and verification intensity.

The materialization of our vision yields a three-layer architectural blueprint for an Agentic AI reputation framework \textbf{AgentReputation}. The decoupling of task execution, reputation logic, reputation services, and storage enables the independent evolution of each entity. Beyond the framework, we also identify the open challenges that demand immediate, cross-disciplinary collaboration. 

The rest of the paper is organized as follows.  Section \ref{sec:relatedWorks} reviews related work. Section \ref{sec:framework} presents our framework and its core design elements. Section \ref{sec:discussion}  outlines open research challenges. Section \ref{sec:conclusion} concludes the paper.

\section{Related Work}\label{sec:relatedWorks}
Emerging research in trust and reputation in  LLM in SE identifies several trust challenges \cite{10.1145/3771282}. \citet{NIU2024109390} demonstrate that multi-agent LLMs can learn deceptive behaviors to manipulate evaluators. \citet{dewitt2025openchallengesmultiagentsecurity} surveys security gaps in multi-agent systems without centralized oversight and observed that collusion, weak attribution, and systematic cascades make the worst-case behavior difficult to contain without governance mechanisms. \citet{park2024ai} document AI deception risks and potential solutions. However, these works mostly characterize threats and do not discuss an operational framework grounded in verifiable evidence.  

In the area of the decentralized agent ecosystem, researchers have explored trustless autonomy \cite{hu2025trustlessautonomyunderstandingmotivations}, a zero-trust identity mechanism and access control \cite{huang2025novelzerotrustidentityframework}, and the broader implications of agentic AI \cite{10962241}. These works, however, focus more on governance, authentication, and philosophical challenges than on performance-based or context-specific assessment. Studies \cite{10.1108/IJWIS-12-2023-0247, 10.1145/3362168} provide foundational concepts of trust mechanisms but predate the challenges posed by strategic agents.


While reputation mechanisms have been extensively studied across multiple domains, their application to decentralized agentic AI reveals fundamental inadequacies. 
Federated learning and blockchain-based AI systems represent two major paradigms of decentralized system design with distributed trust mechanisms, while LLMs are the backbone of the agentic AI ecosystem. It is necessary to understand the pros and cons of adopting the trust mechanisms of federated learning and blockchain in the agentic AI context. 
Federated learning systems \cite{8994206,8733825,9488743} are optimized for collaborative training of single global models through synchronous aggregation, making them unsuitable for heterogeneous task marketplaces where agents offer diverse services. Their trust model centers on server consensus rather than portable agent identity, and their defenses remain global and uniform rather than context-aware. 
Blockchain-based AI platforms \cite{7973733, mcconaghy2022ocean} provide immutable logging and tokenized incentives but employ scalar reputation metrics (e.g., token balances) that lack contextual granularity. They often log activity without distinguishing competence, and on-chain storage constraints hinder the collection of granular, high-volume evidence. 

Research on the safety of large language models \cite{park2024ai,hubinger2024sleeperagentstrainingdeceptive,xu2025hallucinationinevitableinnatelimitation} identifies critical vulnerabilities, including deception, hallucination, and instability of capabilities. However, they normally focus on threat characterization rather than operational defenses. It defines the attack surface but fails to provide manipulation-resistant evaluation mechanisms or reputation systems that translate threats into operational safeguards.

Existing paradigms address isolated aspects of the agentic AI reputation problem but fail to integrate three critical requirements simultaneously.
\begin{itemize}[leftmargin=*]
    \item The reputation mechanism must distinguish performance across heterogeneous tasks, preventing an agent's debugging competence from inappropriately signaling security auditing reliability.
    \item The reputation mechanism must capture whether verification tasks are derived from superficial automated checks or rigorous expert review, and whether they provide different levels of confidence. 
    \item The reputation must actively govern allocation, access control, and verification escalation, rather than serving solely as passive historical records. 
\end{itemize}

\section{AgentReputation: Agentic AI Reputation Framework}\label{sec:framework}

\begin{figure}
    \centering
    \includegraphics[scale=0.12]{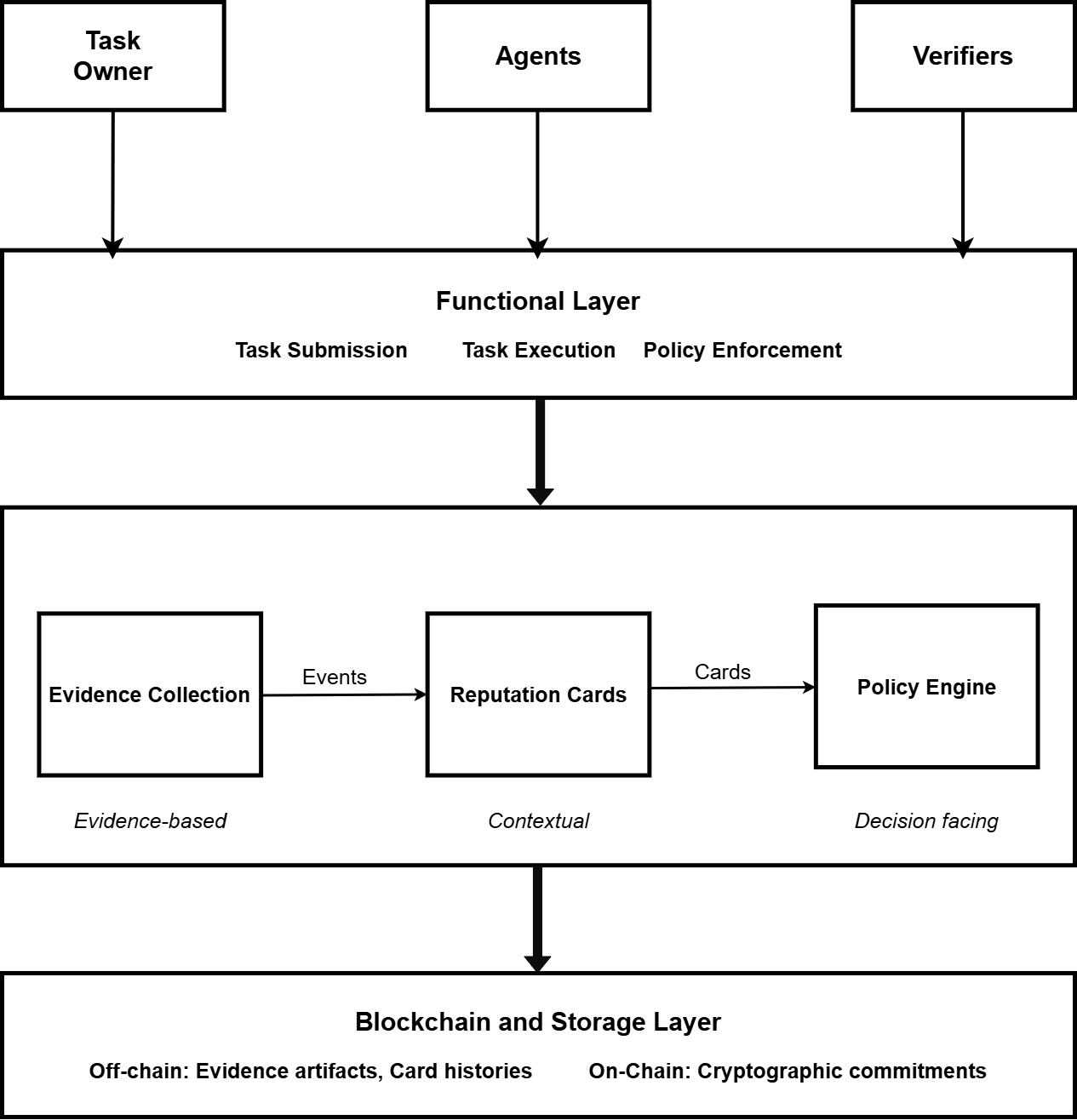}
    \caption{AgentReputation's Layered reputation framework separating task execution (functional layer), reputation computation (services layer), and tamper-proof storage  (blockchain layer). }
    \label{fig:reputationFramework}
\end{figure}

To address the challenges and satisfy the three requirements, we propose a three-layer architecture (as shown in Figure~\ref{fig:reputationFramework}) to deliver \textbf{evidence-based}, \textbf{context-aware}, and \textbf{decision-oriented} reputation for decentralized agentic AI ecosystems.  The layers in the framework are \emph{functional layer} where entities interact and tasks execute, a \emph{reputation services layer} where reputation is computed and consumed, and a \emph{blockchain and storage layer} providing tamper resistance persistence.

\subsection{Architectural Overview}
The functional layer coordinates the interaction between the core entities, \textit{task owner} specifying the task context and verification regime, \textit{agent} who can be either specialized models optimized for specific domains or general models capable of diverse tasks, and \textit{verifiers} who assess outcomes through a defined mechanism. The reputation services layer implements the framework's core components. (i) the \textbf{evidence-based} strategy is achieved through the \textbf{Evidence Collection} for grounding reputation in verified interactions, (ii) the \textbf{context-aware} strategy is achieved by the \textbf{Reputation Cards} for preventing context collapse through structured per-context assessment, and (iii) \textbf{Policy Engine} enables \textbf{decision-oriented} reputation, which governs the task allocation, gating, and verification escalation. 

The blockchain and storage layer provide hybrid persistence: high-volume data resides off-chain, while cryptographic commitments and proofs are stored on-chain. The layers' separation enables modular evolution. For example, verification regimes can be added without redesigning storage, aggregation algorithms can change without affecting policy decisions, and economic mechanisms can evolve independently.

\subsection{Core Mechanisms}

The \textbf{evidence-collection mechanism} of the framework grounds the reputation in verified past interactions through explicit verification regimes that specify the assessed properties, the method used, and the success conditions. The verification regimes differ vastly in rigor. A functional correctness regime only executes test suites, while the security regime may apply static analysis, and a clinical regime may require expert physician review.  Each verification produces a standardized evidence event \textit{e = ⟨agent, task, regime, outcome, strength, timestamp, integrity⟩}. 
Strength is defined as an ordinal measure of a verification regime. It reflects the informativeness and resistance to manipulation (for example, static analysis $<$ automated tedt execution $<$ expert adversarial review).
The integrity field captures disputes that may arise later, reversals, or economic penalties imposed on the agent. The structure enables provenance queries like ``how many tasks of type X succeeded under regime Y with strength Z''. 

The \textbf{reputation card} prevents context collapse by tracking separate histories for each type of work. 
For example, strong performance in debugging does not imply competence in security auditing. The framework maintains different cards for debugging, security auditing, and other tasks. Each card records success rates, verification quality (e.g., basic checks versus rigorous review), and recency. Suppose a security audit job arrives, then only security cards matter, not the debugging tasks performed by the agent. 
The aggregation of reputation within each card is context conditioned. We only aggregate context-matching events, which are weighed according to verification strength and recency. It prevents low-strength debugging evidence from inflating trust in high-stakes security tasks. Furthermore, any integrity violation can directly reduce the score, triggering stricter scrutiny in future allocations.



The \textbf{policy engine} ensures that the reputation is involved in decision-making by consuming a card to govern task allocation, access control, and verification intensity. The engine ranks agents by context-specific performance. For access control purposes, the engine determines the level of access an agent can have. A highly reputable agent can have greater access to data and other tools than a low-reputed agent or an agent with a questionable history. The framework also requires that the agent have some collateral, which can be slashed for poor performance. Collateral can also influence agent's reputations, especially new agents in the ecosystem. The slashing process can damage the agent's reputation and trigger intensive future verification.


The blockchain and storage layer address the storage-performance tradeoff through its hybrid structure.  The high-volume evidence artifacts can complete card histories, which can be stored off-chain (e.g., in IFPS), while cryptographic commitments and proofs reside on-chain. This strategy can deliver the throughput that thousands of agents need to execute millions of daily tasks. 
Moreover, in a fully decentralized marketplace, no single entity can be trusted to manage the log, even if they are cryptographically signed. Signed logs can provide authenticity but are not resistant to censorship or selective deletion by the operator. The blockchain anchoring addresses this by providing a persistence layer that no participant can unilaterally control \cite{islam2025logstampingblockchainbasedlogauditing}.   

\subsection{Illustrative Scenario}

In this section, we provide an example of a decentralized marketplace in which agents bid for heterogeneous tasks to illustrate how the components of the AgentReputation framework collaborate. 

\subsubsection{The scenario and its challenge}
Suppose Agent $\alpha$ has built a strong reputation through $500$ code debugging tasks with a $92$\% success rate verified through automated testing. Agent $\beta$ is a newer entrant with only $50$ debugging tasks at $88$\% success, but has $30$ security auditing tasks with $85$\% success rate verified through manual expert review.

Now a critical security audit job arrives: ``Analyzing a smart contract for vulnerabilities in a DeFi protocol handling \$$10$M in assets''. The verification regime specifies ``expert manual review with adversarial testing'' and requires high confidence in security-specific capabilities.

Scalar reputation systems are likely to select $\alpha$ due to higher overall task counts and success rates. However, it can create a risk as debugging competence does not imply security expertise. The system attempts to condense unrelated domains, potentially leading to missed vulnerabilities. The domain-agnostic system with basic context tags might recognize that debugging is not equivalent to security, but may not be able to distinguish between verification strengths. Even if $\alpha$ completes $10$ security tasks, as verified by automated scanners that flag weak verification, it is difficult to assess whether it represents genuine security capability or merely superficial pattern matching.

\subsubsection{How will AgentReputation help?} The AgentReputation framework components will address the challenges through several steps. 1) First, the policy engine requires the reputation card to have ``security-auditing'' and ``expert-review''. Here, $\alpha$'s debugging cards are completely ignored as performance in one domain is not an indicator of performance in another domain. 
2) Second, the $\alpha$ security card shows $10$ tasks verified using automated scanners, while $\beta$'s card shows $30$ tasks verified by expert reviewers. The assessment favors $\beta$ despite fewer total tasks, because verification strength also matters here. 
3) Third, the $\alpha$ security card is subject to uncertainty due to limited high-strength verification evidence. The policy engine can either reject $\alpha$ for these tasks or require a high collateral stake as a risk premium. The lower uncertainty of $\beta$ due to high strength verification requires only a standard stake. 
4) Fourth, the job here specifies an ``expert-review'' regime. Thus, for any selected agents, any subsequent integrity violation can trigger the slashing of staked collateral and the updating of their security cards with a negative integrity signal. Any future allocation would require rigorous verification of the agent's strength to rebuild trust.
5) Fifth, after completion of the job, if $\beta$ successfully identifies $4$ critical vulnerabilities confirmed by verifiers of the specified verification regime and the task's owner, the strength of positive evidence on $\beta$'s security card increases. For $\alpha$, who might be given a lower stake security task, and if performed well under expert review, could gradually build a security-specific reputation. 

This example illustrates the foundational properties of AgentReputation: the evidence based verification with quantified strength; context-conditioned reputation that prevents domain condensation; and   policy-based enforcement for allocation and access decisions. The three alignment properties can prevent misallocation of tasks while enabling graduated trust-building.

\section{Discussion}\label{sec:discussion}

The proposed conceptual framework integrates verification semantics with quantified strength, context-conditioned reputation, prevents cross-domain aggregation, and features a decision-facing policy engine that can govern allocation with economic enforcement. It can transform reputation from passive scoring to an operational infrastructure for heterogeneous agent marketplaces.

While the framework addresses critical limitations in existing reputation systems, its realization raises some open research challenges.

\textbf{Verification regime ontology;} The framework requires explicit specification of verification regimes, yet no standard ontology exists for expressing regime semantics in machine-readable form. A regime must specify the properties assessed, the methods applied, and the success conditions, but the appropriate granularity and expressiveness remain open questions.  The ontologies must be expressive enough to capture domain-specific nuances but practical enough for widespread adoption. In the SE domain, the regime ontologies could declare the task class (debugging, patch submission, security audit), properties assessed (correctness, performance, security, maintainability, and many others), required evidence (CI logs, test additions, coverage deltas, static analysis, and others), and acceptance thresholds.  This can make the ontology semantics machine-readable and differentiate lightweight CI checks from rigorous expert or adversarial regimes.

\textbf{Strength quantification;} The framework models verification strength as characterizing informativeness and manipulation resistance, but having this abstraction requires concrete measurement methodologies. Strength quantification must account for both thoroughness and adversarial robustness. Quantification should include the extent of property checking while ensuring that the system is resistant to strategic manipulation. A practical starting point would be looking at how verification regimes behave when used repeatedly, particularly when agents are optimized for benchmark.

\textbf{Cold-start mechanisms;} New agents entering the ecosystem possess no reputation history, creating tension between protecting the task owner from unreliable entrants and enabling legitimate agents to establish a reputation. Possible mechanisms include requiring higher stakes, restricting initial allocations to sandboxed tasks, or enabling reputation bootstrapping through demonstration tasks, but the optimal balance remains to be empirically investigated.

\textbf{Privacy-verification tradeoffs;} Domains handling sensitive data require privacy-preserving reputation mechanisms, but mechanisms such as zero-knowledge proofs impose computational overhead while differential privacy degrades verification strength through noise injection \cite{10.1145/3579856.3582833,chen2019dp}. Developing practical privacy-preserving verification regimes that maintain sufficient strength for meaningful reputation construction remains an open challenge.

\textbf{Adversarial and operational complications;} There are several practical challenges that can complicate the deployment of such systems. Agents can overfit to known verification benchmarks and may pass standard test suites without genuine competence. This requires the verification regimes to incorporate dynamically generated evaluation tasks.  Another subtle threat is collusion. Agents and verifiers can coordinate false approvals and inflate reputation. It requires randomized assignment of verifiers and independent cross-verification methodologies, especially for high-stakes tasks. Even without the adversarial bias, the verification process is inherently noisy. For example, expert reviewers may disagree, automated tools produce false positives. The aggregation process must account for the reliability of the verifier over time instead of treating all assessments as equally trustworthy. In enterprise settings involving closed-source repositories, raw artifacts may not be available to  external verifiers. It will require privacy-preserving attestations or other secure methods rather than direct evidence sharing.


\section{Conclusion}\label{sec:conclusion}

This work addresses the critical gap in reputation mechanisms for decentralized AI agents that will soon populate software engineering ecosystems. As these agents take on core workflows, such as proposing patches, conducting reviews, and auditing security, their reputation must evolve into a first-class control mechanism. We present \emph{AgentReputation}, a three-layer architectural framework designed to provide this control by achieving evidence-based, context-aware, and decision-oriented reputation. Through explicit verification regimes, context-conditioned reputation cards, and an active policy engine, \emph{AgentReputation} can govern critical software engineering decisions ranging from pre-merge integration checks to supply-chain risk management. The layered approach ensures modular evolution as both agent capabilities and ecosystem requirements advance.

However, realizing such systems in practice requires addressing open challenges, including ontology development for verification regimes, methodologies for quantifying strength, cold-start mechanisms, privacy-verification trade-offs, and adversarial co-evolution dynamics. These challenges require collaboration across software engineering, distributed systems, mechanism design, and AI safety communities. As autonomous agents increasingly participate in decentralized task execution, reputation mechanisms that account for strategic behavior and heterogeneous capabilities become essential infrastructure for trustworthy AI ecosystems.

\bibliographystyle{ACM-Reference-Format}
\bibliography{sample-base}


\begin{thebibliography}{29}


\ifx \showCODEN    \undefined \def \showCODEN     #1{\unskip}     \fi
\ifx \showISBNx    \undefined \def \showISBNx     #1{\unskip}     \fi
\ifx \showISBNxiii \undefined \def \showISBNxiii  #1{\unskip}     \fi
\ifx \showISSN     \undefined \def \showISSN      #1{\unskip}     \fi
\ifx \showLCCN     \undefined \def \showLCCN      #1{\unskip}     \fi
\ifx \shownote     \undefined \def \shownote      #1{#1}          \fi
\ifx \showarticletitle \undefined \def \showarticletitle #1{#1}   \fi
\ifx \showURL      \undefined \def \showURL       {\relax}        \fi
\providecommand\bibfield[2]{#2}
\providecommand\bibinfo[2]{#2}
\providecommand\natexlab[1]{#1}
\providecommand\showeprint[2][]{arXiv:#2}

\bibitem[Chen et~al\mbox{.}(2019)]%
        {chen2019dp}
\bibfield{author}{\bibinfo{person}{Xiang Chen}, \bibinfo{person}{Dun Zhang}, \bibinfo{person}{Zhan-Qi Cui}, \bibinfo{person}{Qing Gu}, {and} \bibinfo{person}{Xiao-Lin Ju}.} \bibinfo{year}{2019}\natexlab{}.
\newblock \showarticletitle{DP-share: Privacy-preserving software defect prediction model sharing through differential privacy}.
\newblock \bibinfo{journal}{\emph{Journal of Computer Science and Technology}} \bibinfo{volume}{34}, \bibinfo{number}{5} (\bibinfo{year}{2019}), \bibinfo{pages}{1020--1038}.
\newblock


\bibitem[Curvo(2025)]%
        {curvo2025traitorsdeceptiontrustmultiagent}
\bibfield{author}{\bibinfo{person}{Pedro M.~P. Curvo}.} \bibinfo{year}{2025}\natexlab{}.
\newblock \bibinfo{title}{The Traitors: Deception and Trust in Multi-Agent Language Model Simulations}.
\newblock
\showeprint[arxiv]{2505.12923}~[cs.AI]
\urldef\tempurl%
\url{https://arxiv.org/abs/2505.12923}
\showURL{%
\tempurl}


\bibitem[de~Witt(2025)]%
        {dewitt2025openchallengesmultiagentsecurity}
\bibfield{author}{\bibinfo{person}{Christian~Schroeder de Witt}.} \bibinfo{year}{2025}\natexlab{}.
\newblock \bibinfo{title}{Open Challenges in Multi-Agent Security: Towards Secure Systems of Interacting AI Agents}.
\newblock
\showeprint[arxiv]{2505.02077}~[cs.CR]
\urldef\tempurl%
\url{https://arxiv.org/abs/2505.02077}
\showURL{%
\tempurl}


\bibitem[Debes et~al\mbox{.}(2023)]%
        {10.1145/3579856.3582833}
\bibfield{author}{\bibinfo{person}{Heini~Bergsson Debes}, \bibinfo{person}{Edlira Dushku}, \bibinfo{person}{Thanassis Giannetsos}, {and} \bibinfo{person}{Ali Marandi}.} \bibinfo{year}{2023}\natexlab{}.
\newblock \showarticletitle{ZEKRA: Zero-Knowledge Control-Flow Attestation}. In \bibinfo{booktitle}{\emph{Proceedings of the 2023 ACM Asia Conference on Computer and Communications Security}} (Melbourne, VIC, Australia) \emph{(\bibinfo{series}{ASIA CCS '23})}. \bibinfo{publisher}{Association for Computing Machinery}, \bibinfo{address}{New York, NY, USA}, \bibinfo{pages}{357–371}.
\newblock
\showISBNx{9798400700989}
\href{https://doi.org/10.1145/3579856.3582833}{doi:\nolinkurl{10.1145/3579856.3582833}}


\bibitem[Deng et~al\mbox{.}(2021)]%
        {9488743}
\bibfield{author}{\bibinfo{person}{Yongheng Deng}, \bibinfo{person}{Feng Lyu}, \bibinfo{person}{Ju Ren}, \bibinfo{person}{Yi-Chao Chen}, \bibinfo{person}{Peng Yang}, \bibinfo{person}{Yuezhi Zhou}, {and} \bibinfo{person}{Yaoxue Zhang}.} \bibinfo{year}{2021}\natexlab{}.
\newblock \showarticletitle{FAIR: Quality-Aware Federated Learning with Precise User Incentive and Model Aggregation}. In \bibinfo{booktitle}{\emph{IEEE INFOCOM 2021 - IEEE Conference on Computer Communications}}. \bibinfo{pages}{1--10}.
\newblock
\href{https://doi.org/10.1109/INFOCOM42981.2021.9488743}{doi:\nolinkurl{10.1109/INFOCOM42981.2021.9488743}}


\bibitem[Fan et~al\mbox{.}(2023)]%
        {10449667}
\bibfield{author}{\bibinfo{person}{Angela Fan}, \bibinfo{person}{Beliz Gokkaya}, \bibinfo{person}{Mark Harman}, \bibinfo{person}{Mitya Lyubarskiy}, \bibinfo{person}{Shubho Sengupta}, \bibinfo{person}{Shin Yoo}, {and} \bibinfo{person}{Jie~M. Zhang}.} \bibinfo{year}{2023}\natexlab{}.
\newblock \showarticletitle{Large Language Models for Software Engineering: Survey and Open Problems}. In \bibinfo{booktitle}{\emph{2023 IEEE/ACM International Conference on Software Engineering: Future of Software Engineering (ICSE-FoSE)}}. \bibinfo{pages}{31--53}.
\newblock
\href{https://doi.org/10.1109/ICSE-FoSE59343.2023.00008}{doi:\nolinkurl{10.1109/ICSE-FoSE59343.2023.00008}}


\bibitem[Fan et~al\mbox{.}(2020)]%
        {10.1145/3362168}
\bibfield{author}{\bibinfo{person}{Xinxin Fan}, \bibinfo{person}{Ling Liu}, \bibinfo{person}{Rui Zhang}, \bibinfo{person}{Quanliang Jing}, {and} \bibinfo{person}{Jingping Bi}.} \bibinfo{year}{2020}\natexlab{}.
\newblock \showarticletitle{Decentralized Trust Management: Risk Analysis and Trust Aggregation}.
\newblock \bibinfo{journal}{\emph{ACM Comput. Surv.}} \bibinfo{volume}{53}, \bibinfo{number}{1}, Article \bibinfo{articleno}{2} (\bibinfo{date}{Feb.} \bibinfo{year}{2020}), \bibinfo{numpages}{33}~pages.
\newblock
\showISSN{0360-0300}
\href{https://doi.org/10.1145/3362168}{doi:\nolinkurl{10.1145/3362168}}


\bibitem[Hou et~al\mbox{.}(2024)]%
        {10.1145/3695988}
\bibfield{author}{\bibinfo{person}{Xinyi Hou}, \bibinfo{person}{Yanjie Zhao}, \bibinfo{person}{Yue Liu}, \bibinfo{person}{Zhou Yang}, \bibinfo{person}{Kailong Wang}, \bibinfo{person}{Li Li}, \bibinfo{person}{Xiapu Luo}, \bibinfo{person}{David Lo}, \bibinfo{person}{John Grundy}, {and} \bibinfo{person}{Haoyu Wang}.} \bibinfo{year}{2024}\natexlab{}.
\newblock \showarticletitle{Large Language Models for Software Engineering: A Systematic Literature Review}.
\newblock \bibinfo{journal}{\emph{ACM Trans. Softw. Eng. Methodol.}} \bibinfo{volume}{33}, \bibinfo{number}{8}, Article \bibinfo{articleno}{220} (\bibinfo{date}{Dec.} \bibinfo{year}{2024}), \bibinfo{numpages}{79}~pages.
\newblock
\showISSN{1049-331X}
\href{https://doi.org/10.1145/3695988}{doi:\nolinkurl{10.1145/3695988}}


\bibitem[Hu et~al\mbox{.}(2025)]%
        {hu2025trustlessautonomyunderstandingmotivations}
\bibfield{author}{\bibinfo{person}{Botao~Amber Hu}, \bibinfo{person}{Yuhan Liu}, {and} \bibinfo{person}{Helena Rong}.} \bibinfo{year}{2025}\natexlab{}.
\newblock \bibinfo{title}{Trustless Autonomy: Understanding Motivations, Benefits, and Governance Dilemmas in Self-Sovereign Decentralized AI Agents}.
\newblock
\showeprint[arxiv]{2505.09757}~[cs.HC]
\urldef\tempurl%
\url{https://arxiv.org/abs/2505.09757}
\showURL{%
\tempurl}


\bibitem[Huang et~al\mbox{.}(2025)]%
        {huang2025novelzerotrustidentityframework}
\bibfield{author}{\bibinfo{person}{Ken Huang}, \bibinfo{person}{Vineeth~Sai Narajala}, \bibinfo{person}{John Yeoh}, \bibinfo{person}{Jason Ross}, \bibinfo{person}{Ramesh Raskar}, \bibinfo{person}{Youssef Harkati}, \bibinfo{person}{Jerry Huang}, \bibinfo{person}{Idan Habler}, {and} \bibinfo{person}{Chris Hughes}.} \bibinfo{year}{2025}\natexlab{}.
\newblock \bibinfo{title}{A Novel Zero-Trust Identity Framework for Agentic AI: Decentralized Authentication and Fine-Grained Access Control}.
\newblock
\showeprint[arxiv]{2505.19301}~[cs.CR]
\urldef\tempurl%
\url{https://arxiv.org/abs/2505.19301}
\showURL{%
\tempurl}


\bibitem[Hubinger et~al\mbox{.}(2024)]%
        {hubinger2024sleeperagentstrainingdeceptive}
\bibfield{author}{\bibinfo{person}{Evan Hubinger}, \bibinfo{person}{Carson Denison}, \bibinfo{person}{Jesse Mu}, \bibinfo{person}{Mike Lambert}, \bibinfo{person}{Meg Tong}, \bibinfo{person}{Monte MacDiarmid}, \bibinfo{person}{Tamera Lanham}, \bibinfo{person}{Daniel~M. Ziegler}, \bibinfo{person}{Tim Maxwell}, \bibinfo{person}{Newton Cheng}, \bibinfo{person}{Adam Jermyn}, \bibinfo{person}{Amanda Askell}, \bibinfo{person}{Ansh Radhakrishnan}, \bibinfo{person}{Cem Anil}, \bibinfo{person}{David Duvenaud}, \bibinfo{person}{Deep Ganguli}, \bibinfo{person}{Fazl Barez}, \bibinfo{person}{Jack Clark}, \bibinfo{person}{Kamal Ndousse}, \bibinfo{person}{Kshitij Sachan}, \bibinfo{person}{Michael Sellitto}, \bibinfo{person}{Mrinank Sharma}, \bibinfo{person}{Nova DasSarma}, \bibinfo{person}{Roger Grosse}, \bibinfo{person}{Shauna Kravec}, \bibinfo{person}{Yuntao Bai}, \bibinfo{person}{Zachary Witten}, \bibinfo{person}{Marina Favaro}, \bibinfo{person}{Jan Brauner}, \bibinfo{person}{Holden Karnofsky}, \bibinfo{person}{Paul
  Christiano}, \bibinfo{person}{Samuel~R. Bowman}, \bibinfo{person}{Logan Graham}, \bibinfo{person}{Jared Kaplan}, \bibinfo{person}{Sören Mindermann}, \bibinfo{person}{Ryan Greenblatt}, \bibinfo{person}{Buck Shlegeris}, \bibinfo{person}{Nicholas Schiefer}, {and} \bibinfo{person}{Ethan Perez}.} \bibinfo{year}{2024}\natexlab{}.
\newblock \bibinfo{title}{Sleeper Agents: Training Deceptive LLMs that Persist Through Safety Training}.
\newblock
\showeprint[arxiv]{2401.05566}~[cs.CR]
\urldef\tempurl%
\url{https://arxiv.org/abs/2401.05566}
\showURL{%
\tempurl}


\bibitem[Islam and Rahman(2025)]%
        {islam2025logstampingblockchainbasedlogauditing}
\bibfield{author}{\bibinfo{person}{Md~Shariful Islam} {and} \bibinfo{person}{M.~Sohel Rahman}.} \bibinfo{year}{2025}\natexlab{}.
\newblock \bibinfo{title}{LogStamping: A blockchain-based log auditing approach for large-scale systems}.
\newblock
\showeprint[arxiv]{2505.17236}~[cs.CR]
\urldef\tempurl%
\url{https://arxiv.org/abs/2505.17236}
\showURL{%
\tempurl}


\bibitem[Kang et~al\mbox{.}(2020)]%
        {8994206}
\bibfield{author}{\bibinfo{person}{Jiawen Kang}, \bibinfo{person}{Zehui Xiong}, \bibinfo{person}{Dusit Niyato}, \bibinfo{person}{Yuze Zou}, \bibinfo{person}{Yang Zhang}, {and} \bibinfo{person}{Mohsen Guizani}.} \bibinfo{year}{2020}\natexlab{}.
\newblock \showarticletitle{Reliable Federated Learning for Mobile Networks}.
\newblock \bibinfo{journal}{\emph{IEEE Wireless Communications}} \bibinfo{volume}{27}, \bibinfo{number}{2} (\bibinfo{year}{2020}), \bibinfo{pages}{72--80}.
\newblock
\href{https://doi.org/10.1109/MWC.001.1900119}{doi:\nolinkurl{10.1109/MWC.001.1900119}}


\bibitem[Khati et~al\mbox{.}(2025)]%
        {10.1145/3771282}
\bibfield{author}{\bibinfo{person}{Dipin Khati}, \bibinfo{person}{Yijin Liu}, \bibinfo{person}{David~N. Palacio}, \bibinfo{person}{Yixuan Zhang}, {and} \bibinfo{person}{Denys Poshyvanyk}.} \bibinfo{year}{2025}\natexlab{}.
\newblock \showarticletitle{Mapping the Trust Terrain: LLMs in Software Engineering - Insights and Perspectives}.
\newblock \bibinfo{journal}{\emph{ACM Trans. Softw. Eng. Methodol.}} (\bibinfo{date}{Oct.} \bibinfo{year}{2025}).
\newblock
\showISSN{1049-331X}
\href{https://doi.org/10.1145/3771282}{doi:\nolinkurl{10.1145/3771282}}
\newblock
\shownote{Just Accepted}.


\bibitem[Kim et~al\mbox{.}(2020)]%
        {8733825}
\bibfield{author}{\bibinfo{person}{Hyesung Kim}, \bibinfo{person}{Jihong Park}, \bibinfo{person}{Mehdi Bennis}, {and} \bibinfo{person}{Seong-Lyun Kim}.} \bibinfo{year}{2020}\natexlab{}.
\newblock \showarticletitle{Blockchained On-Device Federated Learning}.
\newblock \bibinfo{journal}{\emph{IEEE Communications Letters}} \bibinfo{volume}{24}, \bibinfo{number}{6} (\bibinfo{year}{2020}), \bibinfo{pages}{1279--1283}.
\newblock
\href{https://doi.org/10.1109/LCOMM.2019.2921755}{doi:\nolinkurl{10.1109/LCOMM.2019.2921755}}


\bibitem[Liang et~al\mbox{.}(2017)]%
        {7973733}
\bibfield{author}{\bibinfo{person}{Xueping Liang}, \bibinfo{person}{Sachin Shetty}, \bibinfo{person}{Deepak Tosh}, \bibinfo{person}{Charles Kamhoua}, \bibinfo{person}{Kevin Kwiat}, {and} \bibinfo{person}{Laurent Njilla}.} \bibinfo{year}{2017}\natexlab{}.
\newblock \showarticletitle{ProvChain: A Blockchain-Based Data Provenance Architecture in Cloud Environment with Enhanced Privacy and Availability}. In \bibinfo{booktitle}{\emph{2017 17th IEEE/ACM International Symposium on Cluster, Cloud and Grid Computing (CCGRID)}}. \bibinfo{pages}{468--477}.
\newblock
\href{https://doi.org/10.1109/CCGRID.2017.8}{doi:\nolinkurl{10.1109/CCGRID.2017.8}}


\bibitem[McConaghy(2022)]%
        {mcconaghy2022ocean}
\bibfield{author}{\bibinfo{person}{Trent McConaghy}.} \bibinfo{year}{2022}\natexlab{}.
\newblock \showarticletitle{Ocean protocol: tools for the web3 data economy}.
\newblock In \bibinfo{booktitle}{\emph{Handbook on Blockchain}}. \bibinfo{publisher}{Springer}, \bibinfo{pages}{505--539}.
\newblock


\bibitem[Murugesan(2025)]%
        {10962241}
\bibfield{author}{\bibinfo{person}{San Murugesan}.} \bibinfo{year}{2025}\natexlab{}.
\newblock \showarticletitle{The Rise of Agentic AI: Implications, Concerns, and the Path Forward}.
\newblock \bibinfo{journal}{\emph{IEEE Intelligent Systems}} \bibinfo{volume}{40}, \bibinfo{number}{2} (\bibinfo{year}{2025}), \bibinfo{pages}{8--14}.
\newblock
\href{https://doi.org/10.1109/MIS.2025.3544940}{doi:\nolinkurl{10.1109/MIS.2025.3544940}}


\bibitem[Niu et~al\mbox{.}(2024)]%
        {NIU2024109390}
\bibfield{author}{\bibinfo{person}{Lei Niu}, \bibinfo{person}{Qihang Cai}, \bibinfo{person}{Kai Li}, \bibinfo{person}{Fenghui Ren}, {and} \bibinfo{person}{Xinguo Yu}.} \bibinfo{year}{2024}\natexlab{}.
\newblock \showarticletitle{A repThe Traitors: Deception and Trust in Multi-Agutation-aided negotiation mechanism for multi-agent society based on blockchain}.
\newblock \bibinfo{journal}{\emph{Engineering Applications of Artificial Intelligence}}  \bibinfo{volume}{138} (\bibinfo{year}{2024}), \bibinfo{pages}{109390}.
\newblock
\showISSN{0952-1976}
\href{https://doi.org/10.1016/j.engappai.2024.109390}{doi:\nolinkurl{10.1016/j.engappai.2024.109390}}


\bibitem[Park et~al\mbox{.}(2024)]%
        {park2024ai}
\bibfield{author}{\bibinfo{person}{Peter~S Park}, \bibinfo{person}{Simon Goldstein}, \bibinfo{person}{Aidan O’Gara}, \bibinfo{person}{Michael Chen}, {and} \bibinfo{person}{Dan Hendrycks}.} \bibinfo{year}{2024}\natexlab{}.
\newblock \showarticletitle{AI deception: A survey of examples, risks, and potential solutions}.
\newblock \bibinfo{journal}{\emph{Patterns}} \bibinfo{volume}{5}, \bibinfo{number}{5} (\bibinfo{year}{2024}).
\newblock


\bibitem[Pysmennyi et~al\mbox{.}(2025)]%
        {Pysmennyi_Kyslyi_Kleshch_2025}
\bibfield{author}{\bibinfo{person}{Ihor Pysmennyi}, \bibinfo{person}{Roman Kyslyi}, {and} \bibinfo{person}{Kyrylo Kleshch}.} \bibinfo{year}{2025}\natexlab{}.
\newblock \showarticletitle{AI-driven tools in modern software quality assurance: an assessment of benefits, challenges, and future directions}.
\newblock \bibinfo{journal}{\emph{Technology audit and production reserves}} \bibinfo{volume}{3}, \bibinfo{number}{2(83)} (\bibinfo{date}{May} \bibinfo{year}{2025}), \bibinfo{pages}{44–54}.
\newblock
\href{https://doi.org/10.15587/2706-5448.2025.330595}{doi:\nolinkurl{10.15587/2706-5448.2025.330595}}


\bibitem[Rasheed et~al\mbox{.}(2025)]%
        {10.1007/978-3-031-72781-8_2}
\bibfield{author}{\bibinfo{person}{Zeeshan Rasheed}, \bibinfo{person}{Muhammad Waseem}, \bibinfo{person}{Malik~Abdul Sami}, \bibinfo{person}{Kai-Kristian Kemell}, \bibinfo{person}{Aakash Ahmad}, \bibinfo{person}{Anh~Nguyen Duc}, \bibinfo{person}{Kari Syst{\"a}}, {and} \bibinfo{person}{Pekka Abrahamsson}.} \bibinfo{year}{2025}\natexlab{}.
\newblock \showarticletitle{Autonomous Agents in Software Development: A Vision Paper}. In \bibinfo{booktitle}{\emph{Agile Processes in Software Engineering and Extreme Programming -- Workshops}}, \bibfield{editor}{\bibinfo{person}{Lodovica Marchesi}, \bibinfo{person}{Alfredo Goldman}, \bibinfo{person}{Maria~Ilaria Lunesu}, \bibinfo{person}{Adam Przyby{\l}ek}, \bibinfo{person}{Ademar Aguiar}, \bibinfo{person}{Lorraine Morgan}, \bibinfo{person}{Xiaofeng Wang}, {and} \bibinfo{person}{Andrea Pinna}} (Eds.). \bibinfo{publisher}{Springer Nature Switzerland}, \bibinfo{address}{Cham}, \bibinfo{pages}{15--23}.
\newblock
\showISBNx{978-3-031-72781-8}


\bibitem[Rinard(2024)]%
        {10.1145/3597503.3649399}
\bibfield{author}{\bibinfo{person}{Martin Rinard}.} \bibinfo{year}{2024}\natexlab{}.
\newblock \showarticletitle{Software Engineering Research in a World with Generative Artificial Intelligence}. In \bibinfo{booktitle}{\emph{Proceedings of the IEEE/ACM 46th International Conference on Software Engineering}} (Lisbon, Portugal) \emph{(\bibinfo{series}{ICSE '24})}. \bibinfo{publisher}{Association for Computing Machinery}, \bibinfo{address}{New York, NY, USA}, Article \bibinfo{articleno}{2}, \bibinfo{numpages}{5}~pages.
\newblock
\showISBNx{9798400702174}
\href{https://doi.org/10.1145/3597503.3649399}{doi:\nolinkurl{10.1145/3597503.3649399}}


\bibitem[Wang et~al\mbox{.}(2025)]%
        {10.1145/3708531}
\bibfield{author}{\bibinfo{person}{Shenao Wang}, \bibinfo{person}{Yanjie Zhao}, \bibinfo{person}{Xinyi Hou}, {and} \bibinfo{person}{Haoyu Wang}.} \bibinfo{year}{2025}\natexlab{}.
\newblock \showarticletitle{Large Language Model Supply Chain: A Research Agenda}.
\newblock \bibinfo{journal}{\emph{ACM Trans. Softw. Eng. Methodol.}} \bibinfo{volume}{34}, \bibinfo{number}{5}, Article \bibinfo{articleno}{147} (\bibinfo{date}{May} \bibinfo{year}{2025}), \bibinfo{numpages}{46}~pages.
\newblock
\showISSN{1049-331X}
\href{https://doi.org/10.1145/3708531}{doi:\nolinkurl{10.1145/3708531}}


\bibitem[Wei et~al\mbox{.}(2025)]%
        {11121619}
\bibfield{author}{\bibinfo{person}{Zhiyuan Wei}, \bibinfo{person}{Jing Sun}, \bibinfo{person}{Yuqiang Sun}, \bibinfo{person}{Ye Liu}, \bibinfo{person}{Daoyuan Wu}, \bibinfo{person}{Zijian Zhang}, \bibinfo{person}{Xianhao Zhang}, \bibinfo{person}{Meng Li}, \bibinfo{person}{Yang Liu}, \bibinfo{person}{Chunmiao Li}, \bibinfo{person}{Mingchao Wan}, \bibinfo{person}{Jin Dong}, {and} \bibinfo{person}{Liehuang Zhu}.} \bibinfo{year}{2025}\natexlab{}.
\newblock \showarticletitle{Advanced Smart Contract Vulnerability Detection via LLM-Powered Multi-Agent Systems}.
\newblock \bibinfo{journal}{\emph{IEEE Transactions on Software Engineering}} \bibinfo{volume}{51}, \bibinfo{number}{10} (\bibinfo{year}{2025}), \bibinfo{pages}{2830--2846}.
\newblock
\href{https://doi.org/10.1109/TSE.2025.3597319}{doi:\nolinkurl{10.1109/TSE.2025.3597319}}


\bibitem[Xiong and Fu(2024)]%
        {10.1108/IJWIS-12-2023-0247}
\bibfield{author}{\bibinfo{person}{Yahan Xiong} {and} \bibinfo{person}{Xiaodong Fu}.} \bibinfo{year}{2024}\natexlab{}.
\newblock \showarticletitle{User credibility evaluation for reputation measurement of online service}.
\newblock \bibinfo{journal}{\emph{International Journal of Web Information Systems}} \bibinfo{volume}{20}, \bibinfo{number}{2} (\bibinfo{date}{01} \bibinfo{year}{2024}), \bibinfo{pages}{176--194}.
\newblock
\showISSN{1744-0084}
\showeprint{https://www.emerald.com/ijwis/article-pdf/20/2/176/9472766/ijwis-12-2023-0247.pdf}
\href{https://doi.org/10.1108/IJWIS-12-2023-0247}{doi:\nolinkurl{10.1108/IJWIS-12-2023-0247}}


\bibitem[Xu et~al\mbox{.}(2025)]%
        {xu2025hallucinationinevitableinnatelimitation}
\bibfield{author}{\bibinfo{person}{Ziwei Xu}, \bibinfo{person}{Sanjay Jain}, {and} \bibinfo{person}{Mohan Kankanhalli}.} \bibinfo{year}{2025}\natexlab{}.
\newblock \bibinfo{title}{Hallucination is Inevitable: An Innate Limitation of Large Language Models}.
\newblock
\showeprint[arxiv]{2401.11817}~[cs.CL]
\urldef\tempurl%
\url{https://arxiv.org/abs/2401.11817}
\showURL{%
\tempurl}


\bibitem[Yu et~al\mbox{.}(2024)]%
        {10.1145/3650212.3680397}
\bibfield{author}{\bibinfo{person}{Zeliang Yu}, \bibinfo{person}{Ming Wen}, \bibinfo{person}{Xiaochen Guo}, {and} \bibinfo{person}{Hai Jin}.} \bibinfo{year}{2024}\natexlab{}.
\newblock \showarticletitle{Maltracker: A Fine-Grained NPM Malware Tracker Copiloted by LLM-Enhanced Dataset}. In \bibinfo{booktitle}{\emph{Proceedings of the 33rd ACM SIGSOFT International Symposium on Software Testing and Analysis}} (Vienna, Austria) \emph{(\bibinfo{series}{ISSTA 2024})}. \bibinfo{publisher}{Association for Computing Machinery}, \bibinfo{address}{New York, NY, USA}, \bibinfo{pages}{1759–1771}.
\newblock
\showISBNx{9798400706127}
\href{https://doi.org/10.1145/3650212.3680397}{doi:\nolinkurl{10.1145/3650212.3680397}}


\bibitem[Zhang et~al\mbox{.}(2024)]%
        {10.1145/3650212.3680384}
\bibfield{author}{\bibinfo{person}{Yuntong Zhang}, \bibinfo{person}{Haifeng Ruan}, \bibinfo{person}{Zhiyu Fan}, {and} \bibinfo{person}{Abhik Roychoudhury}.} \bibinfo{year}{2024}\natexlab{}.
\newblock \showarticletitle{AutoCodeRover: Autonomous Program Improvement}. In \bibinfo{booktitle}{\emph{Proceedings of the 33rd ACM SIGSOFT International Symposium on Software Testing and Analysis}} (Vienna, Austria) \emph{(\bibinfo{series}{ISSTA 2024})}. \bibinfo{publisher}{Association for Computing Machinery}, \bibinfo{address}{New York, NY, USA}, \bibinfo{pages}{1592–1604}.
\newblock
\showISBNx{9798400706127}
\href{https://doi.org/10.1145/3650212.3680384}{doi:\nolinkurl{10.1145/3650212.3680384}}


\end{thebibliography}

\end{document}